\documentclass[sn-mathphys]{sn-jnl}

\jyear{2025}%

\theoremstyle{thmstyleone}%
\theoremstyle{thmstyletwo}%
\newtheorem{remark}{Remark}%

\theoremstyle{thmstylethree}%

\usepackage{times}
\usepackage{soul}
\usepackage{url}
\usepackage[utf8]{inputenc}
\usepackage{graphicx}
\usepackage{amsmath}
\usepackage{booktabs}

\usepackage{multicol}
\usepackage{multirow}
\usepackage{color}
\usepackage{xcolor}
\usepackage{booktabs}
\usepackage{graphicx}
\usepackage{nicefrac}
\usepackage{amsmath}
\usepackage{array}
\newcolumntype{P}[1]{>{\centering\arraybackslash}p{#1}}

\usepackage{graphicx}
\usepackage{fancyhdr}
\usepackage{url}
\usepackage{color}
\usepackage{marvosym}

\usepackage{float}

\usepackage{caption}
\usepackage{subcaption}

\usepackage{algorithm}

\usepackage{xfp} 

\begin{document}




\title[Breaking the Euclidean Barrier]{Breaking the Euclidean Barrier: Hyperboloid-Based Biological Sequence Analysis}


\author*[1]{\fnm{Sarwan} \sur{Ali}}\email{sa4559@cumc.columbia.edu}

\author[3]{\fnm{Haris} \sur{Mansoor}}\email{16060061@lums.edu.pk}

\author[2]{\fnm{Murray} \sur{Patterson}}\email{mpatterson30@gsu.edu}

\affil[1]{\orgdiv{Department Of Neurology}, \orgname{Columbia University}, \orgaddress{ \city{New York}, \postcode{10032}, \state{NY}, \country{USA}}}

\affil[2]{\orgdiv{Department Of Computer Science}, \orgname{Georgia State University}, \orgaddress{ \city{Atlanta}, \postcode{30302}, \state{Georgia}, \country{USA}}}

\affil[3]{\orgdiv{Department Of Computer Science}, \orgname{Lahore University of Management Sciences (LUMS)}, \city{Lahore}, \country{Pakistan}}



\abstract{
Genomic sequence analysis plays a crucial role in various scientific and medical domains. Traditional machine-learning approaches often struggle to capture the complex relationships and hierarchical structures of sequence data when working in high-dimensional Euclidean spaces. This limitation hinders accurate sequence classification and similarity measurement.
To address these challenges, this research proposes a method to transform the feature representation of biological sequences into the hyperboloid space. By applying a transformation, the sequences are mapped onto the hyperboloid, preserving their inherent structural information.
Once the sequences are represented in the hyperboloid space, a kernel matrix is computed based on the hyperboloid features. The kernel matrix captures the pairwise similarities between sequences, enabling more effective analysis of biological sequence relationships. This approach leverages the inner product of the hyperboloid feature vectors to measure the similarity between pairs of sequences.
The experimental evaluation of the proposed approach demonstrates its efficacy in capturing important sequence correlations and improving classification accuracy. 
}

\keywords{Bio-sequence Analysis, Sequence Classification, k-mers, Spike Sequence, Feature Vector Representation}

\maketitle

\section{Introduction}
The analysis of biological sequences is a fundamental task in bioinformatics, enabling us to comprehend the structure, function, and evolution of biological molecules~\cite{koonin2002sequence,tayebi2024pseaac2vec}. Recent progress in machine learning and artificial intelligence has paved the way for learning representations of symbolic data, including text, graphs, and multi-relational data~\cite{tillquist2020low, bhattacharjya1999data, cai2018comprehensive, pezeshkpour2018embedding}. In the realm of biological sequence analysis, understanding the characteristics and relationships of sequences is crucial, particularly in genomics, where the examination of genetic or protein sequences provides valuable insights into biological functions~\cite{harder1996canine, nakagawa2018whole}. Nevertheless, conventional machine learning approaches encounter limitations in capturing the intricate relationships and hierarchical structures inherent in genomic sequences due to the high-dimensional Euclidean spaces they operate within.

In the realm of protein sequences, three types of structures are associated with them: Primary, Secondary, and Tertiary Structures~\cite{eisenhaber1995protein}. The primary structure refers to the linear sequence of amino acids comprising the protein chain, while the secondary structure encompasses the local folding patterns and regular repeating structures within the chain. Furthermore, the tertiary structure represents the overall three-dimensional arrangement of the protein molecule.

Hierarchical and tree-like structures are inherent in biological sequences and play a crucial role in understanding the evolutionary processes across species. Phylogenetic trees have been traditionally used to capture these structures and explain the relationships between sequences~\cite{fagan1994p}. However, constructing phylogenetic trees can be computationally intensive and may not be practical for large-scale sequence analysis. Additionally, even if a tree structure is available, calculating Euclidean distance directly from sequence embeddings can result in the loss of important structural information within the biological sequences. Unfortunately, when using Euclidean distance, the preservation of structural information is not achieved.

The adoption of machine learning (ML) models as alternatives to traditional methods has gained popularity due to their ease of use, generalizability, and efficient performance. However, a limitation of many ML models is their reliance on Euclidean distance computations, which leads to the loss of important structural information, such as the Primary, Secondary, and Tertiary Structures, when comparing numerical embeddings of biological sequences.

In recent years, several methods have been proposed for designing embeddings of biological sequences, including popular techniques such as one-hot encoding (OHE)\cite{kuzmin2020machine} and the $k$-mers spectrum\cite{ali2021spike2vec,chourasia2023reads2vec,chourasia2023enhancing,ali2024compression,taslim2023hashing2vec}. OHE-based embeddings suffer from high dimensionality and the loss of order information in amino acids/nucleotides within the sequences~\cite{ali2021k}. On the other hand, the $k$-mers spectrum can preserve the order of amino acids/nucleotides and capture aspects of the primary structure as well as other protein structures~\cite{asim2020k}. However, using Euclidean distance to compare pairs of $k$-mers spectra fails to consider the structural information, limiting the performance of tasks like classification~\cite{ali2024gaussian,ali2024molecular,ali2024preserving,ali2024elliptic}. Furthermore, Euclidean distance often fails to capture the complex and nonlinear relationships inherent in sequence data. Consequently, there is a need for innovative approaches that can provide more effective and meaningful measures for comparing sequences~\cite{ali2025hash,ali2025hist2vec,murad2025sequence}. One alternative to directly working with embeddings is the use of a kernel matrix~\cite{ali2022efficient}, which projects the data into a higher-dimensional space by taking the dot product between the spectra of biological sequences~\cite{ali2022efficient}. However, this approach may not adequately capture all the nonlinear relationships and similarities between data points.

In order to overcome the challenges associated with the complex structures of biological sequences, this study introduces a novel approach that involves transforming the feature representation of these sequences into the hyperboloid space. By applying this transformation, the inherent structural information of the sequences is preserved as they are mapped onto the hyperboloid. The method consists of two steps: first, the conversion of the biological sequences into numerical form using the $k$-mers spectrum, and second, the transformation of the resulting spectra into corresponding hyperboloid feature vectors. This innovative approach shows great potential in capturing the intricate structural properties of biological sequences. Once the sequences are represented in the hyperboloid space, a kernel matrix is computed based on the hyperboloid features. This kernel matrix effectively captures the pairwise similarities between sequences, enabling a more meaningful analysis of the relationships among biological sequences. The approach leverages the inner product of the hyperboloid feature vectors as a measure of similarity between pairs of sequences.

One of the primary advantages of utilizing the hyperboloid space is its ability to preserve the hierarchical and tree-like structures inherent in biological sequences. Unlike Euclidean representations, which often overlook these intricate relationships, the hyperboloid space provides a more accurate representation that captures the complex dependencies among sequences. Furthermore, the proposed method overcomes the limitations of Euclidean spaces by preserving the structural information of the sequences, including their three-dimensional arrangements (tertiary structure) and hidden hierarchical relationships. 

Our contributions in this paper are the following:
\begin{enumerate}
    \item We propose a novel approach that transforms the feature representation of biological sequences into the hyperboloid space, preserving their inherent structural information.
    \item Computation of a kernel matrix based on the hyperboloid features, capturing pairwise similarities between sequences and enabling more effective analysis of biological sequence relationships. 
    \item We also provide theoretical analysis for the designed embeddings.
    \item Experimental evaluation demonstrating the efficacy of the proposed approach in capturing important sequence correlations and improving classification accuracy.
\end{enumerate}

The subsequent sections of this paper are structured as follows: Section~\ref{sec_related_work} presents a comprehensive review of the relevant literature pertaining to the proposed method. The contribution of this paper is outlined and discussed in Section~\ref{sec_proposed_approach}. Details regarding the dataset statistics and experimental setup are provided in Section~\ref{sec_exp_setup}. The results obtained from the proposed and baseline models are presented and analyzed in Section~\ref{sec_results}. Lastly, Section~\ref{sec_conclusion} concludes this study by summarizing the key findings and implications.

\section{Related Work}\label{sec_related_work}
In this section, we provide a comprehensive review of the existing literature and approaches in the field of biological sequence analysis, focusing on distance measures. We discuss the strengths and limitations of traditional methods and identify the specific gaps that our proposed approach aims to overcome.

Traditional distance measures, including Euclidean distance~\cite{ho2004tracing}, Hamming distance~\cite{chappell2017k}, and cosine similarity~\cite{deshpande2002evaluation}, have been widely utilized in biological sequence analysis. These measures assume that sequences can be represented as vectors in a Euclidean space~\cite{kimothi2016distributed,iuchi2021representation,corso2021neural}. While they have demonstrated effectiveness in certain applications, they exhibit limitations when applied to sequences with intricate structures and dependencies~\cite{corso2021neural}. For instance, Euclidean distance fails to capture the hierarchical relationships and tree-like structures that are common in many sequence datasets. Additionally, these measures often encounter challenges when dealing with high-dimensional sequence data and may yield suboptimal results~\cite{liu2003op}.

Graph-based approaches have garnered significant attention in sequence analysis due to their capacity to capture intricate relationships and dependencies. These methods construct a graph where each sequence corresponds to a node, and nodes are connected by edges if they exhibit a certain level of similarity, commonly referred to as a k-nearest neighbor (kNN) graph~\cite{ali2022evaluating}. Additionally, in protein-protein interaction networks, edges can be formed based on the binding interactions between specific proteins~\cite{tsai2009protein}. However, these approaches can be computationally demanding and may encounter difficulties when dealing with large-scale sequence datasets. Furthermore, it remains unclear from the literature whether these methods can effectively capture diverse hierarchical and structural information inherent in biological sequences (e.g., the potential loss of information when constructing a kNN graph based on Euclidean distance).

Embedding-based methods seek to map sequences into low-dimensional vector spaces, where the similarity between biological sequences is reflected in the distances between embeddings~\cite{kimothi2017metric,tayebi2024tcellr2vec, tayebi2023t}. Approaches such as $k$-mers spectrum~\cite{ali2021k,ali2021spike2vec,tayebi2021robust} have demonstrated promising performance in supervised and unsupervised tasks. However, these methods often suffer from computational complexity or face challenges in preserving the hierarchical and structural relationships inherent in sequences.

Generating a kernel or gram matrix has become a popular approach for calculating the pairwise similarity between sequences, offering an alternative to traditional embedding methods~\cite{ali2021k}. However, the computational cost associated with computing pairwise similarities can be significant. To mitigate this challenge, researchers in~\cite{ali2022efficient} proposed an approximate method that enhances the efficiency of kernel computation. Their approach involves evaluating the dot product between the spectra of two sequences. The resulting kernel matrix can then be employed as input for kernel classifiers, such as Support Vector Machines (SVM), or non-kernel classifiers utilizing kernel PCA~\cite{hoffmann2007kernel}, for classification tasks~\cite{ali2021k}. Despite these advancements, these methods still encounter difficulties in capturing the underlying hierarchical and structural information intrinsic to biological sequences.

In recent years, there has been growing interest in leveraging hyperbolic geometry for sequence analysis~\cite{corso2021neural,nickel2017poincare}. Hyperbolic space possesses unique properties that make it particularly suitable for capturing hierarchical and tree-like structures~\cite{corso2021neural}. Previous studies have demonstrated the effectiveness of hyperbolic representations in various tasks, including text classification~\cite{chen2020hyperbolic}, network analysis~\cite{park2021unsupervised}, and knowledge graph embeddings~\cite{sala2018representation}. However, the application of hyperbolic geometry to biological sequence analysis remains relatively unexplored, highlighting the need for specialized distance measures that harness the advantages of hyperbolic space for the classification of biological sequences.

\begin{figure}[h!]
    \centering
    \includegraphics[scale=0.4]{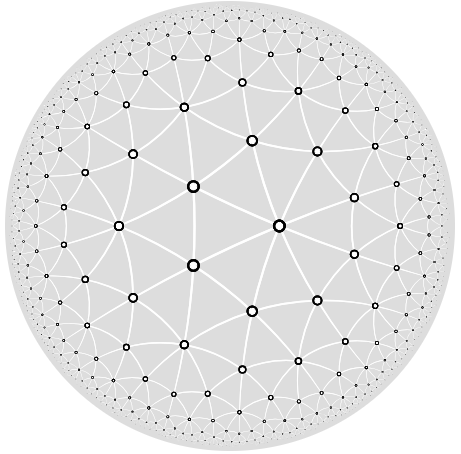}
    \caption{Hyperboloid isosceles tiling, each triangle has two equal sides and equal angles, we can observe negative curvature and exponential growth.}
    \label{fig:1}
\end{figure}

\section{Proposed Approach}\label{sec_proposed_approach}


Given an \textcolor{black}{vector space} $\mathscr{X} \in \mathbb{R}^{n+1}$ (e.g., proteins, etc.), function : $\mathscr{X} \times \mathscr{X} \to \mathbb{R}$ is called a
kernel function. Kernel functions are used to quantify the similarity between a pair of objects $X, Y \in \mathscr{X}$. 
We propose a kernel function in  hyperbolic geometry. Hyperbolic geometry is a non-Euclidean geometry that studies space of constant negative curvature. The underlying hyperbolic geometry allows us to learn similarity better than Euclidean space.

\paragraph{\textbf{Input Embedding Representation}}
In this study, we employ the $k$-mers spectrum~\cite{ali2021k} as the input for our model. The $k$-mers spectrum represents sequences by capturing the frequency of all possible subsequences of length $k$, where $k>0$. It provides a concise encoding of sequences by incorporating the frequencies of all possible $k$-mers. Compared to other encoding schemes like one-hot encoding (OHE)~\cite{kuzmin2020machine}, the $k$-mers spectrum offers several advantages. OHE suffers from high dimensionality and sparsity, particularly for large alphabets or long sequences. In contrast, the $k$-mers spectrum preserves positional and contextual information within the sequence. It takes into account the occurrence and frequencies of subsequences, enabling the detection of patterns, motifs, and dependencies within the sequence, which are crucial for various sequence analysis tasks.

\begin{remark}
We also experimented with the one-hot encoding-based representation of biological sequences as input for the proposed methods. However, the results were not promising compared to the $k$-mers-based spectrum, primarily due to the reasons discussed above. Therefore, we have not included those results in this paper.
\end{remark}

We divide our method into two steps: we first compute the pairwise distance between the data points in hyperbolic space. In the second step, we use kernel PCA to get the final embeddings, which are then used for the underlying supervised analysis.

\subsection{Hyperboloid Model}

In geometry, the hyperboloid model is a model of n-dimensional geometry in which points are represented on the forward sheet $S+$ of a two-sheeted hyperboloid in ($n+1$)-dimensional Minkowski space or by the displacement vectors from the origin to those points. \textcolor{black}{In the hyperboloid model, $m$-dimensional subspaces ($m$-planes) are formed by intersecting the forward sheet $S^+$ with ($m+1$)-dimensional hyperplanes in Minkowski space that pass through the origin. Alternatively, these can be constructed using the exterior (wedge) product of $m$ displacement vectors from the origin.} 
Similar to how spherical distance is inherited from Euclidean distance when the $n$-sphere is embedded in ($n+1$)-dimensional Euclidean space, hyperbolic space is embedded isometrically in Minkowski space. The same ideas (such as points, lines, angles, and circles) can be applied to hyperbolic geometry when they are transferred to the forward sheet of a hyperboloid. 
Representing data in hyperbolic space has an intrinsic advantage as many natural processes have a structure better suited for hyperbolic space. This is because many phenomena have an exponential relationship, which becomes evident in hyperbolic space.

Figure \ref{fig:1} shows a visualization of a particular type of tiling (isosceles) in hyperbolic geometry. In a hyperbolic isosceles tiling, the tiles used are hyperbolic isosceles triangles: they have two equal sides and angles. These triangles are created by cutting a hyperbolic plane into congruent triangles, each having two sides of the same length.





Let $X=(x_0,x_1 \cdots x_n)$ and $Y=(y_0,y_1 \cdots y_n)$ be two points on the positive hyperbolic sheet, the distance between $X$ and $Y$ is given by: 
\begin{equation}\label{eq:1}
d(X,Y)=cosh^{-1}(B(X,Y))
\end{equation}
where, 
\begin{equation}\label{eq:2}
\textcolor{black}{B(X,Y)=x_0 y_0 - x_1 y_1 - x_2y_2 - \cdots - x_ny_n}
\end{equation}
\textcolor{black}{
The domain of Equation~\ref{eq:1} is $B(X,Y) \geq 1$, where $cosh^{-1}(1) = 0$ (corresponding to identical points) and $cosh^{-1}(B(X,Y)) > 0$ for $B(X,Y) > 1$.
} 
The expression $d(X,Y)$ can serve as a kernel function in hyperbolic space. A kernel function typically satisfies the following two properties: non-negativity, and symmetry. A kernel with these properties will loosely have the interpretation as a similarity quantification; that is $d(X,Y)$  represents a similarity metric on input space.
\begin{itemize}
    \item \bf{Non-Negativity:} $d(X,Y)>0$
    \item \bf{Symmetry:} $d(X,Y)=d(Y,X)$
\end{itemize}

\textcolor{black}{
The expression \( B(X, Y) = x_0 y_0 - \sum_{i=1}^n x_i y_i \) represents the Lorentzian inner product \( \langle X, Y \rangle_L \) in Minkowski space satisfying \( \langle X, X \rangle_L = x_0^2 - \sum_{i=1}^n x_i^2 = 1 \) (with \( x_0 > 0 \)) and similarly for \( Y \). \\
\textbf{Proof:} \\
Let \( p = \|\vec{x}\| = \sqrt{\sum_{i=1}^n x_i^2} \), so \( x_0 = \sqrt{1 + p^2} \). Similarly, let \( q = \|\vec{y}\| \), so \( y_0 = \sqrt{1 + q^2} \).
Then,
\[
B(X, Y) = x_0 y_0 - \vec{x} \cdot \vec{y} \geq x_0 y_0 - \|\vec{x}\| \|\vec{y}\| = \sqrt{1 + p^2} \sqrt{1 + q^2} - p q,
\]
where the inequality holds by the (standard) Cauchy-Schwarz inequality on the spatial Euclidean vectors \( \vec{x} \) and \( \vec{y} \), with equality when \( \vec{x} \) and \( \vec{y} \) are parallel.
It remains to show that the lower bound \( f(p, q) = \sqrt{1 + p^2} \sqrt{1 + q^2} - p q \geq 1 \). Rationalize the expression:
\[
f(p, q) = \frac{(1 + p^2)(1 + q^2) - (p q)^2}{\sqrt{1 + p^2} \sqrt{1 + q^2} + p q} = \frac{1 + p^2 + q^2}{\sqrt{1 + p^2} \sqrt{1 + q^2} + p q}.
\]
To verify \( f(p, q) \geq 1 \): \\
- At \( p = q = 0 \), \( f(0, 0) = 1 \). \\
- For \( p = q \), \( f(p, p) = \frac{1 + 2p^2}{1 + p^2 + p^2} = \frac{1 + 2p^2}{1 + 2p^2} = 1 \). \\
- For unequal values, e.g., \( q = 0 \), \( f(p, 0) = \sqrt{1 + p^2} > 1 \) for \( p > 0 \). \\
The minimum value of 1 occurs when \( p = q \) and the spatial vectors are aligned (corresponding to \( X = Y \)); otherwise, \( B(X, Y) > 1 \). This confirms the domain condition for \( \cosh^{-1}(B(X, Y)) \).
}

since,
\begin{equation}\label{eq:3}
cosh^{-1}(z) = ln\left(z+\sqrt{z^2-1}\right) \ \ \ \   1 \leq z < \infty
\end{equation}
Using the Taylor series expansion of $ln$
\begin{equation}\label{eq:4}
ln(z) = (z-1)-\frac{(z-1)^2}{2}+\frac{(z-1)^3}{3}-\frac{(z-1)^4}{4}+ \dots
\end{equation}
Taylor series expansion of $cosh^{-1}(z)$ around $1$ is given below
\begin{equation}\label{eq:5}
\begin{split}
cosh^{-1}(z) = (z-1+\sqrt{z^2-1}) \\ -\frac{(z-1+\sqrt{z^2-1})^2}{2}+
\frac{(z-1+\sqrt{z^2-1})^3}{3} \\ -\frac{(z-1+\sqrt{z^2-1})^4}{4}  \dots
\end{split}
\end{equation}
Using binomial expansion the above expression can be simplified into
\begin{equation}\label{eq:9}
\begin{split}
cosh^{-1}(z) = (2z-1-\frac{1}{2z}-\frac{1}{8z^3}-\frac{1}{32z^5} \cdots)- \\
\frac{1}{2}(2z-1-\frac{1}{2z}-\frac{1}{8z^3}-\frac{1}{32z^5} \cdots)^2+ \\
\frac{1}{3}(2z-1-\frac{1}{2z}-\frac{1}{8z^3}-\frac{1}{32z^5} \cdots)^3 + \cdots
\end{split}
\end{equation}
Thus $d(X,Y)$ as a kernel map inputs to infinite dimensional polynomial feature space with both negative and positive exponents. \textcolor{black}{The above expansions demonstrate that $d(X, Y)$ can be expressed as an infinite series resembling a polynomial kernel, allowing it to map inputs implicitly to a high-dimensional feature space. This means that the hyperboloid kernel effectively corresponds to a polynomial kernel of infinite degree. Polynomial kernels are well known for their ability to approximate complex nonlinear decision boundaries: \\ - Low-degree terms capture simple linear and quadratic patterns. \\
- High-degree terms represent increasingly intricate interactions between input features.
}

\begin{figure*}[h!]
    \centering
    \includegraphics[scale=0.28]{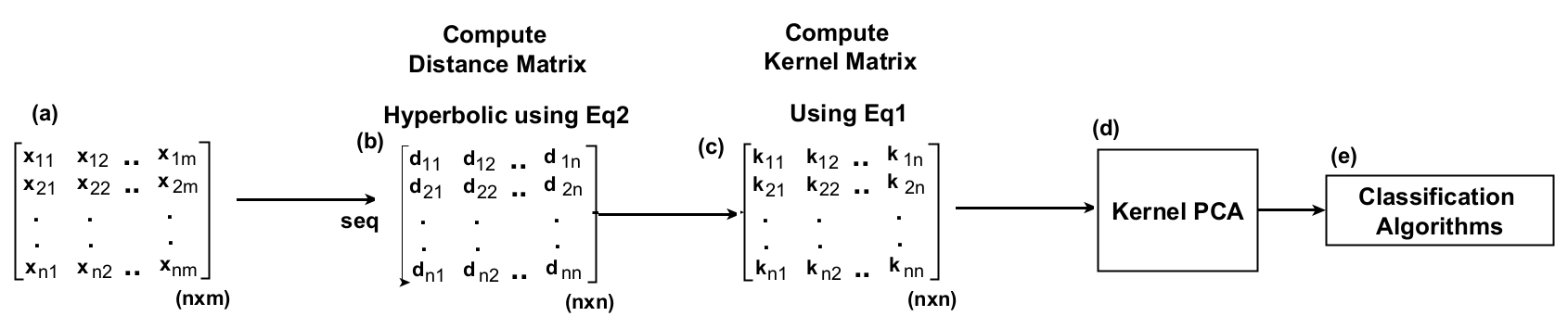}
    \caption{Workflow for Hyperboliod Distance-based classification. Each row of the input matrix (a) represents a sequence vector computed using $k$-mers spectrum. The sequence is first converted into kernel matrix using hyperboloid distance (b) and (c), then the kernel matrix is processed through Kernel-PCA (d) and used in the classification algorithms.}
    \label{process_flow_chart}
\end{figure*}

The pseudocode for computing the Hyperboloid kernel is given in Algorithm~\ref{algo_1} and the flow chart is provided in Figure~\ref{process_flow_chart}. The algorithm takes a set of biological sequences as input and computes $k$-mers spectrum-based numerical embeddings (line 1). It then iterates for all pairs of embeddings (upper triangle only due to symmetry property) and computes the Hyperboloid distance using Equation~\ref{eq:1} and Equation~\ref{eq:2} (line 7). The kernel matrix is further processed to kernel-PCA and then used in classification algorithms.

\subsection{Theoretical Proof}\label{sec:PSD}

Mercer's theorem, as discussed by~\cite{xu2019generalized}, offers a fundamental underpinning for the utilization of kernel methods in machine learning. The theorem establishes that any positive semi-definite kernel function can be represented as the inner product of two functions in a feature space of high dimensionality. This enables us to bypass the explicit computation of the high-dimensional feature space and instead operate directly with the kernel function in the input space. Meeting the conditions of Mercer's theorem allows for the application of kernel methods like support vector machines (SVMs) and kernel principal component analysis (KPCA). These methods bring forth advantages such as efficient computation, the flexibility to model intricate relationships between data, and the ability to handle non-linearly separable data. Additionally, the desirable properties of positive definiteness and full rank, which will be discussed in detail below, offer further benefits such as facilitating the construction of a reproducible kernel Hilbert Space (RKHS). This framework provides a basis for a unique and reproducible Kernel.

A kernel function $d(X_i, X_j)\  \forall \ X_i,x_j \in \mathbb{R}^n$ is called a Mercer kernel if it satisfies the following properties:

\begin{enumerate}
    \item $d(X_i, X_j)$ is symmetric, i.e. $d(X_i, X_j) = d(X_j, X_i)$ $\forall$ $X_i$ and $X_j$.
    \item For any set of data points ${X_1, X_2, ..., X_n}$, the kernel matrix $D$ defined by $D_{ij} = d(X_i, X_j)$ is positive semi-definite.
   
\end{enumerate}

To prove this, let's consider the following conditions: 

\paragraph{\textbf{Symmetry: }}
The Hyperboloid kernel function is symmetric since, 
\begin{equation} \label{eq:5}
d(X_i,X_j)=d(X_j,X_i)
 \end{equation}

\textcolor{black}{
The Hyperboloid kernel function is symmetric because:
\begin{align}
B(X,Y) &= x_0y_0 - \sum_{i=1}^{n} x_iy_i \\
&= y_0x_0 - \sum_{i=1}^{n} y_ix_i = B(Y,X)
\end{align}
Since $B(X,Y) = B(Y,X)$, it immediately follows that:
\begin{equation}
d(X,Y) = \cosh^{-1}(B(X,Y)) = \cosh^{-1}(B(Y,X)) = d(Y,X)
\end{equation}.}

\paragraph{\textbf{Postitive Semi Definite (PSD) Analysis: }}

\textcolor{black}{The kernel matrix $\mathbf{D} \in \mathbb{R}^{n \times n}$ is defined as $D_{ij} = d(X_i, X_j)$ for all data points $X_1, X_2, \ldots, X_n$ in our dataset. To verify positive semi-definiteness (PSD), we must show that for any vector $\mathbf{v} \in \mathbb{R}^n$, the quadratic form $\mathbf{v}^T\mathbf{D}\mathbf{v} \geq 0$.
\begin{equation}\label{eq:8}
\mathbf{v}^T\mathbf{D}\mathbf{v} = \sum_{i=1}^{n}\sum_{j=1}^{n} v_i v_j D_{ij} = \sum_{i=1}^{n}\sum_{j=1}^{n} v_i v_j d(X_i, X_j)
\end{equation}
Equation \ref{eq:8} follows directly from the definition of matrix multiplication:
\begin{equation}\label{eq:9}
\mathbf{v}^T\mathbf{D}\mathbf{v} = \sum_{i=1}^{n} v_i^2 d(X_i, X_i) + \sum_{i,j=1; i\neq j}^{n} v_i v_j d(X_i, X_j)
\end{equation}
Equation \ref{eq:9} separates the diagonal terms ($i=j$) from the off-diagonal terms ($i \neq j$):
\begin{equation}
\sum_{i=1}^{n} v_i^2 d(X_i, X_i) \geq -\sum_{i,j=1; i\neq j}^{n} v_i v_j d(X_i, X_j)
\end{equation}
We can add a positive number in $d(X_i, X_i)$ such that the inequality in the above equation is always satisfied. This has no effect on machine learning algorithms as these algorithms utilize the pairwise kernel measure for $i \neq j$. Thus, adding a positive number in $d(X_i, X_i)$ will make the kernel positive semidefinite without affecting the results.} 

Since the Hyperboloid function is a Mercer kernel, there exists a function  $\phi$ that maps $X_i, X_j$ into another space (possibly with much higher dimensions) such that 
\begin{equation}\label{eq:10}
d(X_i,X_j)=\phi(X_i)^T\phi(X_j)
\end{equation}

where, we can use $d$ as a kernel since we know $\phi$ exists, even if we do not know what $\phi$ is. We can use $\phi(X_i)^T\phi(X_j)$ as a comparison metric between samples. In a number of machine learning and statistical models, the feature vector $X$ only enters the model through comparisons with other feature vectors. If we kernelize these inner products, When we are allowed to compute only $d(X_i, X_j)$  in computation and avoid working in feature space, we work in an $n \times n$ space (the Kernel matrix).

\begin{algorithm}[h!]
\caption{Hyperboloid Kernel Matrix}
\label{algo_1}
\begin{algorithmic}[1]
\Statex \textbf{Input: } Set of Biological Sequences $n$
\Statex \textbf{Output: } Hyperboloid kernel matrix K
\State embed $\gets$ \Call{kmersSpectrum}{n} \Comment{Using method from~\cite{ali2021k}}
\State K $\gets$ np.zeros($\vert n \vert$, $\vert n \vert$)

\State Initialize kernel matrix $K$ with zeros
\For{$i$ in range(n)} \Comment{for all embeddings $i$}
    \For{$j$ in range(n)} \Comment{for all embeddings $j$}
        \If{$i \leq j$}
            \State kernel $\gets$ $cosh^{-1}($embed[i], embed[j]) \Comment{using Eq.~\ref{eq:1} and Eq.~\ref{eq:2}}
            \State Set $K[i, j]$ = kernel
            \State Set $K[j, i] = K[i, j]$
        \EndIf
    \EndFor
\EndFor
\State  \textbf{Return} $K$
\end{algorithmic}
\end{algorithm}

\subsection{Kernel PCA~\cite{hoffmann2007kernel}}\label{sec_kernel_PCA}
Once the kernel matrix is generated, we can utilize kernel PCA (Principal Component Analysis) to obtain a low-dimensional embedding of the sequences. PCA is a dimensionality reduction technique that finds a set of orthogonal axes (principal components) along which the data exhibits the most variation. Kernel PCA extends this method to nonlinear data by using the kernel matrix to implicitly map the data into a high-dimensional feature space.

In kernel PCA, the principal components are computed based on the eigenvectors of the kernel matrix rather than the original data. This allows for the discovery of patterns and structures in the data. The resulting low-dimensional embeddings, also known as kernel principal components, capture the essential information while preserving the nonlinear relationships among the sequences.

By applying kernel PCA to the kernel matrix generated from the hyperboloid distance, we can obtain a reduced-dimensional representation of the sequences. This representation can be used for various downstream tasks, such as clustering, classification, or visualization, where the nonlinear structure of the data is important.

The advantage of using kernel PCA with the Hyperboloid distances lies in its ability to capture the underlying nonlinear relationships among the sequences. The incorporation of hyperbolic geometry in distance calculations allows for a more accurate representation of the data, particularly when dealing with hierarchical or tree-like structures.

Overall, the proposed approach combines the Hyperboloid model with kernel PCA to construct a kernel matrix and derive low-dimensional embeddings. This allows for the exploration and analysis of sequence data in hyperbolic space, facilitating the discovery of hierarchical relationships and capturing nonlinear patterns that may not be fully captured by traditional Euclidean methods.

\subsection{Justification of Using Kernel Matrix}\label{app_justification}
Generating a kernel matrix from the Hyperboloid distance offers several technical justifications and benefits:

\begin{enumerate}
    \item \textbf{Nonlinearity:} The Hyperboloid distance is a measure that accounts for the underlying hyperbolic geometry, representing a non-Euclidean space. Through the utilization of this distance, a kernel matrix can be constructed, facilitating the implicit mapping of data into a higher-dimensional feature space. This mapping enables the capture of intricate and nonlinear relationships, which proves advantageous in scenarios involving hierarchical or tree-like structures commonly found in biological sequences.    
    \item \textbf{Flexibility and Generalization:} The kernel matrix obtained from the distance measure based on the Hyperboloid can be effectively employed with diverse machine learning algorithms designed for kernel matrices. This flexibility enables the application of various techniques, including kernel PCA, kernel SVM, and more. By leveraging these algorithms, we can harness the rich representation capabilities of the kernel matrix to tackle various tasks, such as dimensionality reduction and classification.
    \item \textbf{Preserving Nonlinear Information:} Kernel PCA applied to the kernel matrix effectively captures the intrinsic nonlinear characteristics of the data, enabling the extraction of low-dimensional embeddings that preserve the underlying structure and patterns. Through the projection of the data onto the principal components, we can retain the discriminative information while reducing the dimensionality. This approach proves valuable for tasks such as data visualization and handling high-dimensional datasets.
\end{enumerate}

\section{Experimental Setup}\label{sec_exp_setup}
This section outlines the experimental setup, encompassing the utilized dataset, evaluation metrics, and implementation specifics. We present the experimental results, conduct a comparative analysis between the proposed approach and existing methods, and offer a comprehensive interpretation of the findings.
All experiments were performed using Python on a system with a Core i5 processor operating at a frequency of 2.4 GHz, 32 GB of memory, and the Windows 10 operating system.

We utilize three real-world biological sequence datasets, consisting of both protein and nucleotide sequences. Table~\ref{tbl_data_statistics} provides a summary of the datasets used in our experiments.

\begin{table}[h!]
\centering
\resizebox{0.9\textwidth}{!}{
\begin{tabular}{p{1.4cm}cccccp{1.5cm}p{6cm}}
\toprule
\multirow{2}{1.2cm}{Name} & \multirow{2}{*}{$\vert$ Seq. $\vert$} & \multirow{2}{*}{$\vert$ Classes $\vert$} & \multicolumn{3}{c}{Sequence Statistics} & \multirow{2}{*}{Reference} & \multirow{2}{*}{Description} \\
\cmidrule{4-6}
& & & Max & Min & Mean & & \\
\midrule
\multirow{3}{1.2cm}{Spike7k} & \multirow{3}{*}{7000} & \multirow{3}{*}{22} & \multirow{3}{*}{1274} & \multirow{3}{*}{1274} & \multirow{3}{*}{1274.00} & \multirow{3}{1.5cm}{~\cite{gisaid_website_url}} & The dataset contains spike protein sequences of the SARS-CoV-2 virus along with information about the coronavirus lineages associated with each sequence. \\
\midrule
\multirow{2}{1.2cm}{Human DNA} & \multirow{2}{*}{4380} & \multirow{2}{*}{7} & \multirow{2}{*}{18921} & \multirow{2}{*}{5} & \multirow{2}{*}{1263.59} & \multirow{2}{*}{~\cite{human_dna_website_url}} & The dataset consists of unaligned nucleotide sequences used for classifying the gene family to which humans belong. \\
\midrule
\multirow{4}{1.2cm}{Coronavirus Host} & \multirow{4}{*}{5558} & \multirow{4}{*}{21} & \multirow{4}{*}{1584} & \multirow{4}{*}{9} & \multirow{4}{*}{1272.36} & \multirow{4}{1.5cm}{ViPR~\cite{pickett2012vipr}, GISAID~\cite{gisaid_website_url}} & The dataset includes spike protein sequences belonging to various clades of the Coronaviridae family, accompanied by labels indicating the infected host (e.g., Humans, Bats, Chickens, etc.). \\
\bottomrule
\end{tabular}
}
\caption{Summary of dataset statistics for the three datasets used in our evaluation.}
\label{tbl_data_statistics}
\end{table}

We have carefully selected a range of recently proposed methods as baselines from various categories of embedding generation. These categories include feature engineering, traditional kernel matrix generation (with kernel PCA), neural networks, pre-trained language models, and pre-trained transformers specifically designed for protein sequences. Detailed information about these baseline models can be found in Table~\ref{tbl_sota_detail}.

\begin{table}[h!]
\centering
\resizebox{0.9\textwidth}{!}{
\begin{tabular}{p{2cm}p{2cm}p{8.5cm}c}
\toprule
Method & Category & Description & Source \\
\midrule \midrule
\multirow{2}{*}{PWM2Vec} & \multirow{2}{2cm}{Feature Engineering} & \multirow{2}{8.5cm}{This method takes a biological sequence as input and generates fixed-length numerical embeddings.} & \multirow{2}{*}{~\cite{ali2022pwm2vec}} \\
& \\
\cmidrule{2-4}
\multirow{3}{*}{String Kernel} & \multirow{3}{2cm}{Kernel Matrix} & This approach designs an $n \times n$ kernel matrix that can be used with kernel classifiers or kernel PCA to obtain feature vectors based on principal components. & \multirow{4}{*}{~\cite{ali2022efficient}} \\
\cmidrule{2-4}
\multirow{3}{*}{SeqVec} & \multirow{2}{2cm}{Pretrained Language Model} & This method takes biological sequences as input and fine-tunes the weights based on a pre-trained model to obtain the final embeddings. & \multirow{3}{*}{~\cite{heinzinger2019modeling}} \\
\cmidrule{2-4}
\multirow{2}{*}{ProteinBERT} & \multirow{2}{2cm}{Pretrained Transformer} & This is a pretrained protein sequence model that utilizes the Transformer/Bert architecture to classify the given biological sequences. & \multirow{3}{*}{~\cite{brandes2022proteinbert}} \\
\cmidrule{2-4}
\multirow{2}{*}{WDGRL} & \multirow{4}{2cm}{Neural Network (NN)} & \multirow{4}{8cm}{This method takes the one-hot representation of a biological sequence as input and generates embeddings using a neural network by minimizing a loss function.} & \multirow{2}{*}{~\cite{shen2018wasserstein}} \\
& \\
\multirow{2}{*}{AutoEncoder} & & & \multirow{2}{*}{~\cite{xie2016unsupervised}} \\
& \\
\bottomrule
\end{tabular}
}
\caption{\textcolor{black}{Description of different baseline models.}}
\label{tbl_sota_detail}
\end{table}

To assess the performance of various models, we employ a range of evaluation metrics, including average accuracy, precision, recall, weighted F1 score, macro F1 score, Receiver Operator Characteristic Curve (ROC AUC), and training runtime. For metrics specifically designed for binary classification, we adopt the one-vs-rest approach for multi-class classification.
\textcolor{black}{
The supervised analysis involves the utilization of both linear and non-linear classifiers. Specifically, we employ: Support Vector Machine (SVM)~\cite{cortes1995support}, which finds optimal separating hyperplanes by maximizing the margin between classes; Naive Bayes (NB)~\cite{rish2001empirical}, a probabilistic classifier based on Bayes' theorem with independence assumptions; Multi-Layer Perceptron (MLP)~\cite{gardner1998artificial}, a feedforward neural network with multiple hidden layers; K-Nearest Neighbors (KNN)~\cite{cover1967nearest}, an instance-based method that classifies based on majority voting of k nearest neighbors; Random Forest (RF)~\cite{breiman2001random}, an ensemble method using multiple decision trees; Logistic Regression (LR)~\cite{hosmer2000applied}, a linear classifier using the logistic function; and Decision Tree (DT)~\cite{quinlan1986induction}, a tree-based rule classifier.
For our method, the kernel matrix generated from hyperboloid distances is first processed through kernel PCA to obtain low-dimensional embeddings (as described in Section~\ref{sec_kernel_PCA}). These embeddings are then used as input features to each of the seven classifiers listed above. All classifiers were implemented using scikit-learn~\cite{pedregosa2011scikit} with default parameters unless otherwise specified.
}
To visually evaluate if the embeddings generated from the Hyperboloid Model using kernel PCA keep similar points closer to each other in the hyperbolic geometry space, we use the t-Distributed Stochastic Neighbor Embedding (t-SNE) method~\cite{van2008visualizing}.

\subsection{Data Visualization}
\begin{figure}[h!]
\centering
\begin{subfigure}{.33\textwidth}
  \centering
  \includegraphics[scale=0.08]{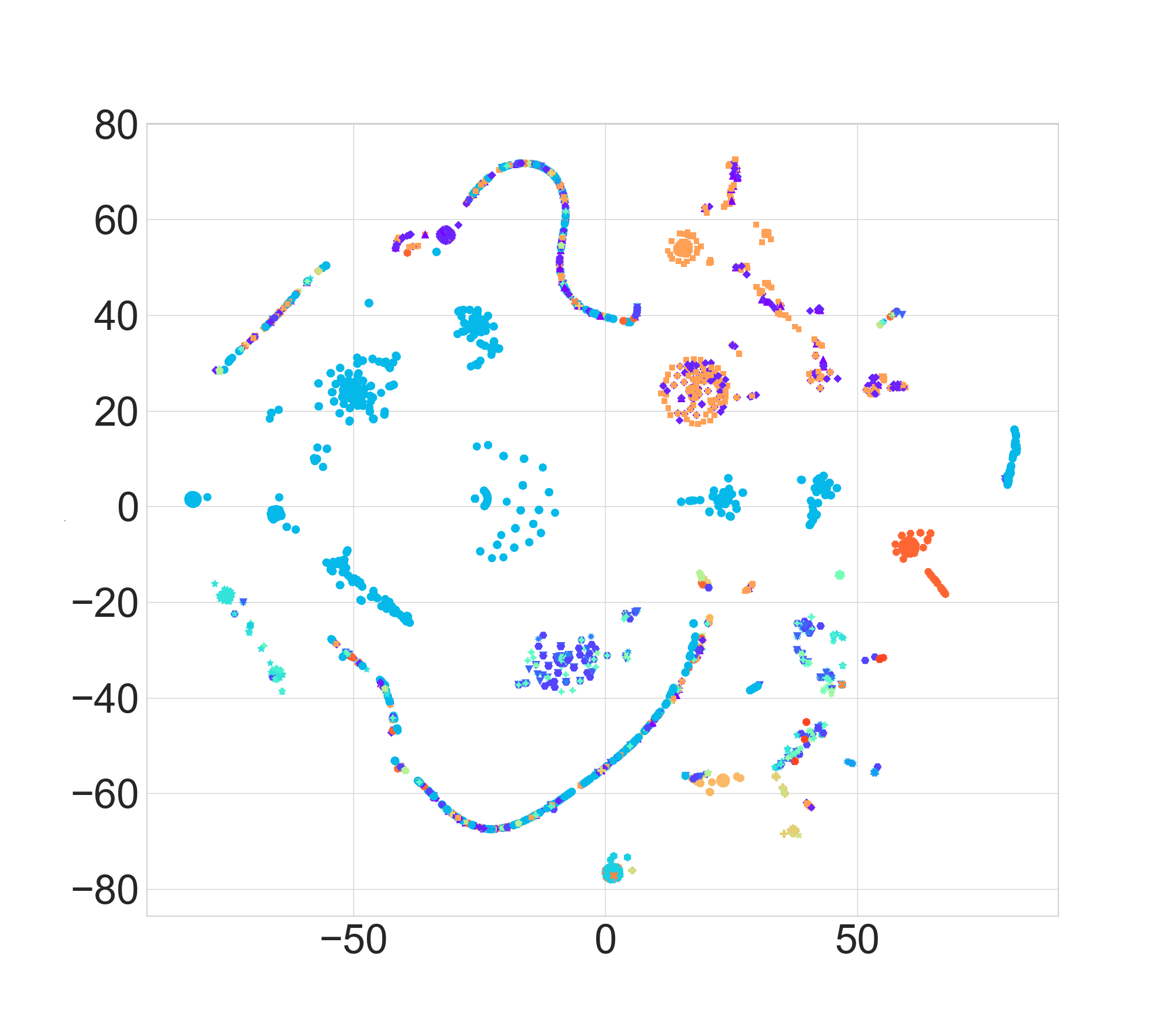}
  \caption{Spike7k}
\end{subfigure}%
\begin{subfigure}{.33\textwidth}
  \centering
  \includegraphics[scale=0.08]{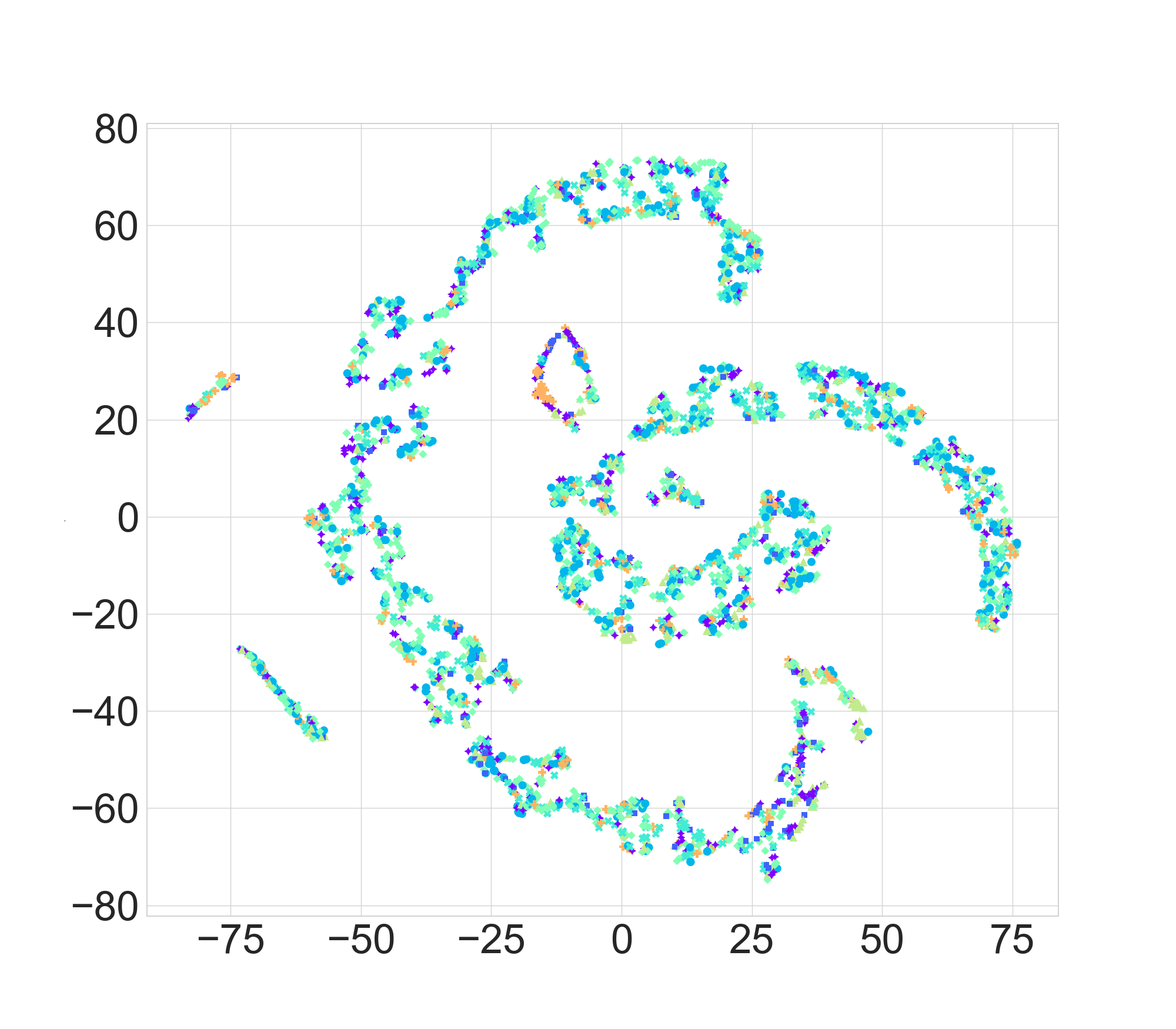}
  \caption{Human DNA}
\end{subfigure}%
\begin{subfigure}{.33\textwidth}
  \centering
  \includegraphics[scale=0.08]{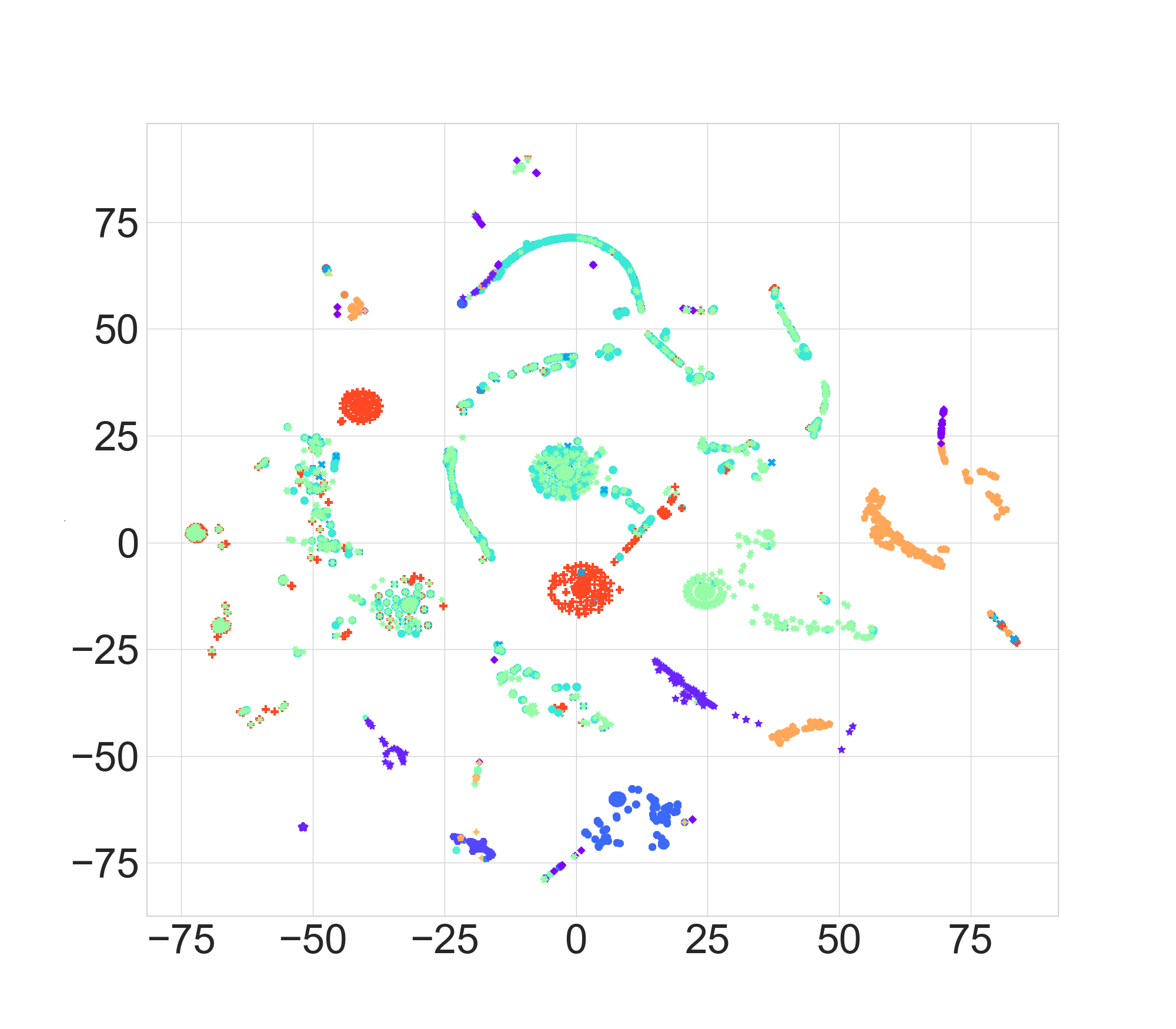}
  \caption{Coronavirus Host}
\end{subfigure}%
\caption{t-SNE plots for Hyperboloid Model using Spike7k, Human DNA, and Coronavirus Host datasets.}
\label{fig_all_tsne}
\end{figure}

To visually evaluate if the embeddings generated from the Hyperboloid Model using kernel PCA keep similar points closer to each other in the hyperbolic geometry space, we use the t-Distributed Stochastic Neighbor Embedding (t-SNE) method~\cite{van2008visualizing} to get 2-D representations for the embeddings and plot them using the scatterplot. 
Including t-SNE in this study enables the assessment of the proposed techniques through visual representations of biological sequence classification and grouping, based on their similarities. This visualization is highly valuable in evaluating the effectiveness of the proposed methods in capturing the underlying patterns and relationships within the dataset.
The tSNE plots for Spike7k, Human DNA, and Coronavirus Host dataset are shown in Figure~\ref{fig_all_tsne}. 
For the Spike7k dataset, we can observe that the majority class (i.e., B.1.1.7) and other classes like P.1 and B.1.526 are nicely grouped together. Similarly, for the Human DNA dataset, although we can observe some overlapping of data points belonging to different classes, the classes like Transcription Factor and Synthetase show several smaller groups. Moreover, for the Coronavirus Host dataset, we can observe clear grouping for the classes such as Weasel, Environment, and Turtle. These groupings highlights the fact that the proposed embedding is able to preserve similar classes together in low dimensional representation.

\section{Results And Discussion}\label{sec_results}
Table~\ref{tbl_results_host_classification} presents the average classification results for the coronavirus host dataset, specifically focusing on spike proteins. Notably, the proposed hyperboloid model-based method outperforms all other methods in terms of average accuracy, precision, recall, and weighted F1 scores. These findings highlight the superior performance of our model in accurately classifying the spike proteins. However, it is worth mentioning that the baseline PWM2Vec method achieves the highest macro F1 macro and ROC-AUC score among all the methods considered while our method achieves comparable performance in this case.

\begin{table}[h!]
\centering
\resizebox{0.9\textwidth}{!}{
 \begin{tabular}{@{\extracolsep{6pt}}p{1.5cm}lp{1.1cm}p{1.1cm}p{1.1cm}p{1.3cm}p{1.3cm}p{1.1cm}p{1.7cm}}
    \toprule
        \multirow{2}{*}{Embeddings} & \multirow{2}{*}{Algo.} & \multirow{2}{*}{Acc. $\uparrow$} & \multirow{2}{*}{Prec. $\uparrow$} & \multirow{2}{*}{Recall $\uparrow$} & \multirow{2}{1.7cm}{F1 (Weig.) $\uparrow$} & \multirow{2}{1.7cm}{F1 (Macro) $\uparrow$} & \multirow{2}{1.2cm}{ROC AUC $\uparrow$} & Train Time (sec.) $\downarrow$\\
        \midrule \midrule	
        \multirow{7}{1.2cm}{PWM2Vec}
        & SVM & 0.799 & 0.806 & 0.799 & 0.801 & 0.648 & 0.859 & 44.793	\\
        & NB & 0.381 & 0.584 & 0.381 & 0.358 & 0.400 & 0.683 & \underline{2.494}    \\
        & MLP & 0.782 & 0.792 & 0.782 & 0.778 & 0.693 & 0.848 & 21.191  \\
        & KNN & 0.786 & 0.782 & 0.786 & 0.779 & 0.679 & 0.838 & 12.933  \\
        & RF & \underline{0.836} & \underline{0.839} & \underline{0.836} & \underline{0.828} & \underline{\textbf{0.739}} & \underline{\textbf{0.862}} & 7.690    \\
        & LR & 0.809 & 0.815 & 0.809 & 0.800 & 0.728 & 0.852 & 274.91  \\
        & DT & 0.801 & 0.802 & 0.801 & 0.797 & 0.633 & 0.829 & 4.537    \\

        \cmidrule{2-9}
        
        \multirow{7}{1.5cm}{String Kernel}
            & SVM & 0.601 & 0.673 & 0.601 & 0.602 & 0.325 & 0.624 & 5.198 \\
             & NB & 0.230 & 0.665 & 0.230 & 0.295 & 0.162 & 0.625 & \underline{0.131} \\
             & MLP & 0.647 & 0.696 & 0.647 & 0.641 & 0.302 & 0.628 & 42.322 \\
             & KNN & 0.613 & 0.623 & 0.613 & 0.612 & 0.310 & 0.629 & 0.434 \\
             & RF & \underline{0.668} & 0.692 & \underline{0.668} & \underline{0.663} & \underline{0.360} & \underline{0.658} & 4.541 \\
             & LR & 0.554 & \underline{0.724} & 0.554 & 0.505 & 0.193 & 0.568 & 5.096 \\
             & DT &  0.646 & 0.674 & 0.646 & 0.643 & 0.345 & 0.653 & 1.561 \\
                  \cmidrule{2-9}
                  
        \multirow{7}{1.9cm}{WDGRL}
                 & SVM & 0.329 & 0.108 & 0.329 & 0.163 & 0.029 & \underline{0.500} & 2.859 \\
         & NB & 0.004 & 0.095 & 0.004 & 0.007 & 0.002 &  0.496 & \textbf{\underline{0.008}}  \\
         & MLP & 0.328 & 0.136 & 0.328 & 0.170 & 0.032 & 0.499 & 5.905  \\
         & KNN & 0.235 & 0.198 & 0.235 & 0.211 & \underline{0.058} & 0.499 & 0.081  \\
         & RF & 0.261 & 0.196 & 0.261 & \underline{0.216} & 0.051 & 0.499 &  1.288 \\
         & LR & \underline{0.332} & 0.149 & \underline{0.332} & 0.177 & 0.034 & \underline{0.500} & 0.365   \\
         & DT & 0.237 & \underline{0.202} & 0.237 & 0.211 & 0.054 & 0.498 & 0.026  \\ 
          \cmidrule{2-9}
        \multirow{7}{1.5cm}{Autoencoder}
         & SVM & 0.602 & 0.588 & 0.602 & 0.590 &  0.519 & 0.759 & 2575.9 \\
         & NB & 0.261 & 0.520 & 0.261 & 0.303 & 0.294 & 0.673 & 21.74 \\
         & MLP & 0.486 & 0.459 & 0.486 & 0.458 & 0.216 &  0.594 & 29.93 \\
         & KNN & 0.763 & 0.764 & 0.763 & 0.755 & 0.547 & 0.784 & \underline{18.51} \\
         & RF &  \underline{0.800} & \underline{0.796} & \underline{0.800} & \underline{0.791} & \underline{0.648} & \underline{0.815} &  57.90\\
         & LR & 0.717 & 0.750 & 0.717 & 0.702 & 0.564 & 0.812 & 11072.6 \\
         & DT & 0.772 & 0.767 & 0.772 & 0.765 & 0.571 & 0.808 & 121.36 \\ 
          \cmidrule{2-9}
        \multirow{7}{1.5cm}{SeqVec}
         & SVM & 0.711 & 0.745 & 0.711 & 0.698 & 0.497 & 0.747 & 0.751 \\
         & NB & 0.503 & 0.636 & 0.503 & 0.554 & 0.413 & 0.648 & \underline{0.012} \\
         & MLP & 0.718 & 0.748 & 0.718 & 0.708 & 0.407 & 0.706 & 10.191  \\
         & KNN & 0.815 & 0.806 & 0.815 & 0.809 & 0.588 & 0.800 & 0.418  \\
         & RF & \underline{0.833} & \underline{0.824} & \underline{0.833} & \underline{0.828} & \underline{0.678} & \underline{0.839} &  1.753   \\
         & LR & 0.673 & 0.683 & 0.673 & 0.654 & 0.332 & 0.660 & 1.177   \\
         & DT & 0.778 & 0.786 & 0.778 & 0.781 & 0.618 & 0.825 & 0.160  \\ 
                   \cmidrule{2-9}
        \multirow{1}{1.9cm}{Protein Bert}
         & \_ & 0.799 &  0.806 & 0.799 & 0.789 & 0.715 & 0.841 & 15742.9 \\
          \cmidrule{2-9} 
           \multirow{7}{1.5cm}{Hyperboloid Model (ours)}
              & SVM & 0.346 & 0.261 & 0.346 & 0.207 & 0.040 & 0.504 & 13.423 \\
             & NB & 0.585 & 0.633 & 0.585 & 0.585 & 0.460 & 0.741 & \underline{0.288} \\
             & MLP & 0.751 & 0.751 & 0.751 & 0.746 & 0.474 & 0.726 & 7.102 \\
             & KNN & \textbf{0.877} & \textbf{0.870} & \textbf{0.877} & \textbf{0.873} & 0.622 & \underline{0.832} & 1.204 \\
             & RF & 0.833 & 0.833 & 0.833 & 0.824 & \underline{0.655} & 0.810 & 5.127 \\
             & LR & 0.351 & 0.257 & 0.351 & 0.214 & 0.042 & 0.505 & 2.261 \\
             & DT & 0.782 & 0.780 & 0.782 & 0.777 & 0.556 & 0.790 & 1.315 \\

         \bottomrule
         \end{tabular}
    }
    \caption{Classification results (averaged over $5$ runs) for different evaluation metrics for \textbf{Coronavirus Host Dataset}. The best values are shown in bold. The best value for each embedding method is shown with the underline.}
    \label{tbl_results_host_classification}
\end{table}

Table~\ref{tbl_results_classification_spike7k} presents the classification results (averaged over 5 runs) for the proposed hyperboloid-based method, as well as the baseline models, on the Spike7k dataset. It is evident that the proposed methods consistently outperform the baseline models in terms of predictive performance. These methods consistently achieve higher scores across various evaluation metrics, demonstrating their superiority compared to other embedding methods.

\begin{table}[h!]
\centering
\resizebox{0.9\textwidth}{!}{
 \begin{tabular}{@{\extracolsep{6pt}}p{1.5cm}lp{1.1cm}p{1.1cm}p{1.1cm}p{1.3cm}p{1.3cm}p{1.1cm}p{1.7cm}}
    \toprule
        \multirow{2}{*}{Embeddings} & \multirow{2}{*}{Algo.} & \multirow{2}{*}{Acc. $\uparrow$} & \multirow{2}{*}{Prec. $\uparrow$} & \multirow{2}{*}{Recall $\uparrow$} & \multirow{2}{1.7cm}{F1 (Weig.) $\uparrow$} & \multirow{2}{1.7cm}{F1 (Macro) $\uparrow$} & \multirow{2}{1.2cm}{ROC AUC $\uparrow$} & Train Time (sec.) $\downarrow$\\
        \midrule \midrule
        \multirow{7}{1.2cm}{PWM2Vec}
        & SVM & 0.818 & 0.820 & 0.818 & 0.810 & 0.606 & \underline{0.807} & 22.710 \\
        & NB & 0.610 & 0.667 & 0.610 & 0.607 & 0.218 & 0.631 & 1.456 \\ 
        & MLP & 0.812 & 0.792 & 0.812 & 0.794 & 0.530 & 0.770 & 35.197  \\ 
        & KNN & 0.767 & 0.790 & 0.767 & 0.760 & 0.565 & 0.773 & \underline{1.033} \\
        & RF & \underline{0.824} & \underline{0.843} & \underline{0.824} & \underline{0.813} & \underline{0.616} & 0.803 & 8.290 \\
        & LR & 0.822 & 0.813 & 0.822 & 0.811 & 0.605 & 0.802 & 471.659 \\ 
        & DT & 0.803 & 0.800 & 0.803 & 0.795 & 0.581 & 0.791 & 4.100 \\
        \cmidrule{2-9}
        \multirow{7}{1.9cm}{String Kernel}
        & SVM  & 0.845 & 0.833 & \underline{0.846} & 0.821 & 0.631 & 0.812 & 7.350 \\
        & NB   & 0.753 & 0.821 & 0.755 & 0.774 & 0.602 & 0.825 & \underline{0.178} \\
        & MLP  & 0.831 & 0.829 & 0.838 & 0.823 & 0.624 & 0.818 & 12.652 \\
        & KNN  & 0.829 & 0.822 & 0.827 & 0.827 & 0.623 & 0.791 & 0.326 \\
        & RF   & \underline{0.847} & \underline{0.844} & 0.841 & \underline{0.835} & \underline{0.666} & 0.824 & 1.464 \\
        & LR   & 0.845 & 0.843 & 0.843 & 0.826 & 0.628 & 0.812 & 1.869 \\
        & DT   & 0.822 & 0.829 & 0.824 & 0.829 & 0.631 & \underline{0.826} & 0.243 \\
        \cmidrule{2-9}
        \multirow{7}{1.2cm}{WDGRL}  
        & SVM & 0.792 & 0.769 & 0.792 & 0.772 & 0.455 & 0.736 & 0.335 \\
        & NB & 0.724 & 0.755 & 0.724 & 0.726 & 0.434 & 0.727 & 0.018 \\
        & MLP & 0.799 & 0.779 & 0.799 & 0.784 & 0.505 & 0.755 & 7.348 \\
        & KNN & \underline{0.800} & \underline{0.799} & \underline{0.800} & \underline{0.792} & 0.546 & 0.766 & 0.094 \\
        & RF & 0.796 & 0.793 & 0.796 & 0.789 & \underline{0.560} & \underline{0.776} & 0.393 \\
        & LR & 0.752 & 0.693 & 0.752 & 0.716 & 0.262 & 0.648 & 0.091 \\
        & DT & 0.790 & \underline{0.799} & 0.790 & 0.788 & 0.557 & 0.768 & \underline{\textbf{0.009}} \\
        \cmidrule{2-9}
        \multirow{7}{1.5cm}{Autoencoder}
        & SVM &  0.699 & 0.720 & 0.699 & 0.678 & 0.243 & 0.627 & 4018.02 \\
        & NB & 0.490 & 0.533 & 0.490 & 0.481 & 0.123 & 0.620 & 24.6372 \\
        & MLP & 0.663 & 0.633 & 0.663 & 0.632 & 0.161 & 0.589 & 87.4913 \\
        & KNN & 0.782 & 0.791 & 0.782 &  0.776 & 0.535 & 0.761 & \underline{24.5597} \\
        & RF & \underline{0.814} & \underline{0.803} & \underline{0.814} & \underline{0.802} & \underline{0.593} & \underline{0.793} &  46.583\\
        & LR & 0.761 & 0.755 & 0.761 & 0.735 & 0.408 & 0.705 & 11769.02 \\
        & DT & 0.803 & 0.792 & 0.803 & 0.792 & 0.546 & 0.779 & 102.185 \\ 
        \cmidrule{2-9}
        \multirow{7}{1.5cm}{SeqVec}
        & SVM & \underline{0.796} & 0.768 & \underline{0.796} & 0.770 & 0.479 & 0.747 & 1.0996 \\
        & NB & 0.686 & 0.703 & 0.686 & 0.686 & 0.351 & 0.694 & \underline{0.0146} \\
        & MLP & \underline{0.796} & 0.771 & \underline{0.796} & 0.771 & 0.510 & 0.762 & 13.172 \\   
        & KNN & 0.790 & 0.787 & 0.790 & \underline{0.786} & \underline{0.561} & 0.768 &  0.6463 \\ 
        & RF & 0.793 & \underline{0.788} & 0.793 & \underline{0.786} & 0.557 & \underline{0.769} &  1.8241\\
        & LR & 0.785 & 0.763 & 0.785 & 0.761 & 0.459 & 0.740 & 1.7535 \\
        & DT & 0.757 & 0.756 & 0.757 & 0.755 & 0.521 & 0.760 & 0.1308 \\ 
        \cmidrule{2-9}
        \multirow{1}{1.5cm}{Protein Bert}
         & \_ &  0.836 & 0.828 & 0.836 & 0.814 & 0.570 & 0.792 & 14163.52 \\
        \cmidrule{2-9}
            \multirow{7}{1.5cm}{Hyperboloid Model (ours)}
             & SVM & 0.721 & 0.660 & 0.721 & 0.680 & 0.172 & 0.591 & 4.215 \\
             & NB &  0.597 & 0.783 & 0.597 & 0.663 & 0.531 & 0.769 & 0.406 \\
             & MLP & 0.751 & 0.753 & 0.751 & 0.747 & 0.534 & 0.756 & 4.896 \\
             & KNN & \textbf{0.849} & \textbf{0.847} & \textbf{0.849} & \textbf{0.838} & \textbf{0.672} & 0.824 & \underline{0.382} \\
             & RF &  0.824 & 0.821 & 0.824 & 0.807 & 0.625 & 0.791 & 5.508 \\
             & LR &  0.679 & 0.496 & 0.679 & 0.571 & 0.139 & 0.563 & 3.648 \\
             & DT &  0.830 & 0.832 & 0.830 & 0.825 & 0.636 & \textbf{0.835} & 1.384 \\
         \bottomrule
         \end{tabular}
}
 \caption{Classification results (averaged over $5$ runs) on \textbf{Spike7k} dataset for different evaluation metrics. The best values are shown in bold. The best value for each embedding method is shown with the underline.}
    \label{tbl_results_classification_spike7k}
\end{table}

\begin{figure}[h!]
\centering
\begin{subfigure}{.25\textwidth}
  \centering
  \includegraphics[scale = 0.120] {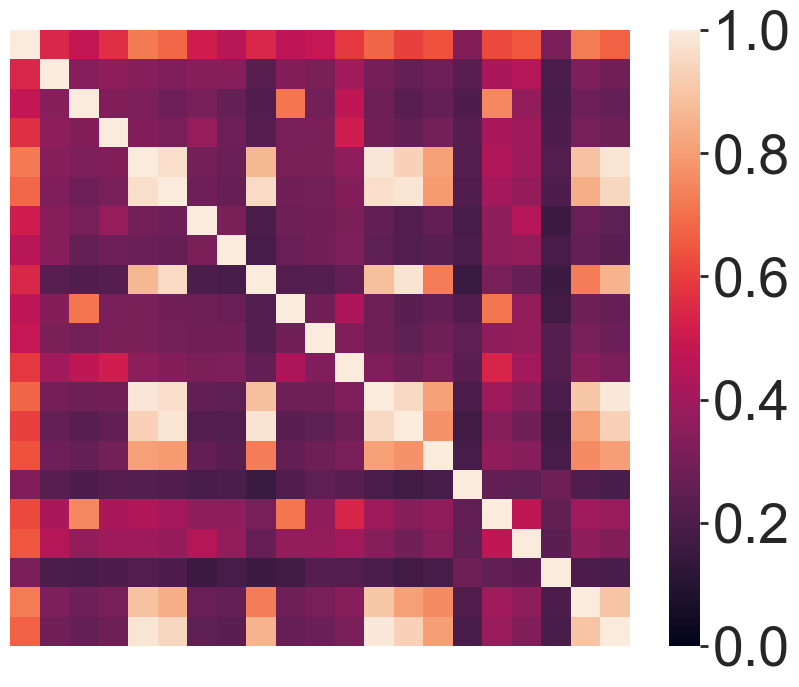}
  \caption{Spike2Vec}
\end{subfigure}%
\begin{subfigure}{.25\textwidth}
  \centering
  \includegraphics[scale = 0.120] {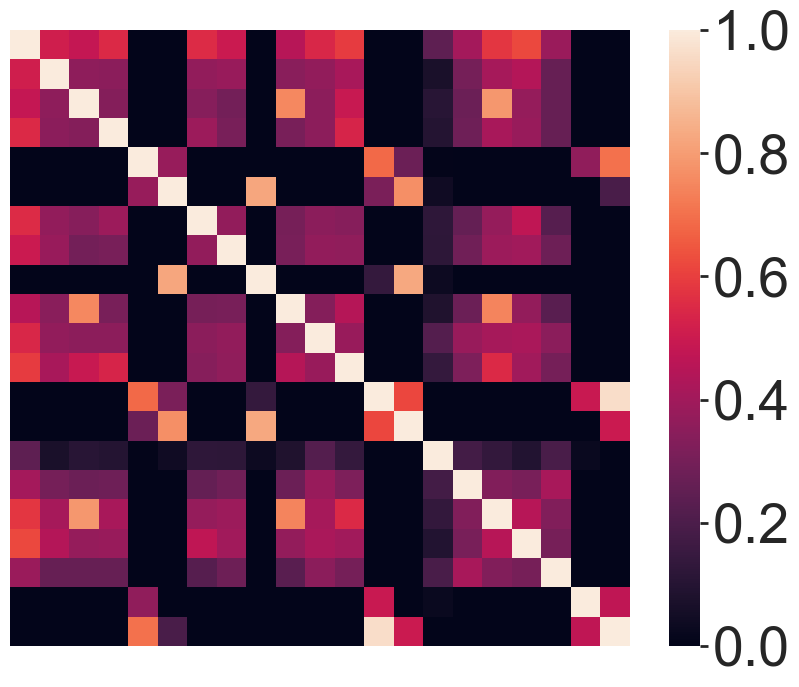}
  \caption{Hyperboloid Model}
\end{subfigure}
\caption{Heatmap for classes in \textbf{Coronavirus Host dataset}. 
}
\label{fig_heat_map_5558}
\end{figure}

\begin{figure}[h!]
\centering
\begin{subfigure}{.25\textwidth}
  \centering
  \includegraphics[scale = 0.120] {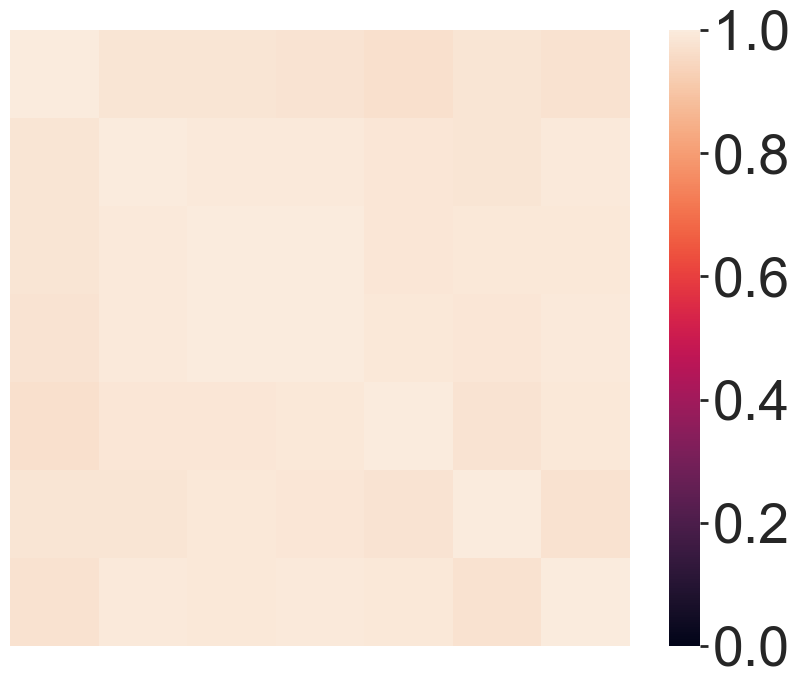}
  \caption{Spike2Vec}
\end{subfigure}%
\begin{subfigure}{.25\textwidth}
  \centering
  \includegraphics[scale = 0.120] {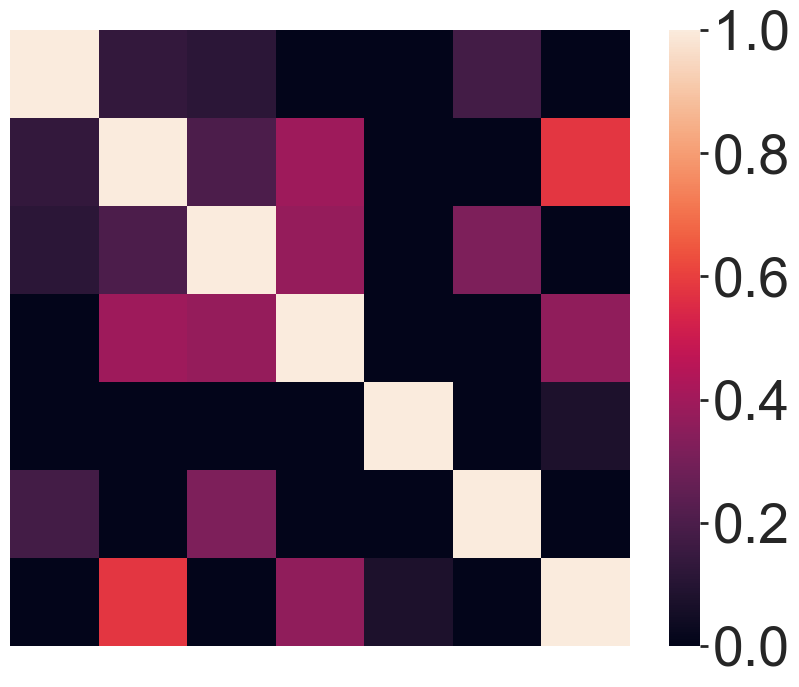}
  \caption{Hyperboloid Model}
\end{subfigure}
\caption{Heatmap for classes in \textbf{Human DNA dataset}. 
}
\label{fig_heat_map_4380}
\end{figure}

\begin{figure}[h!]
\centering
\begin{subfigure}{.25\textwidth}
  \centering
  \includegraphics[scale = 0.120] {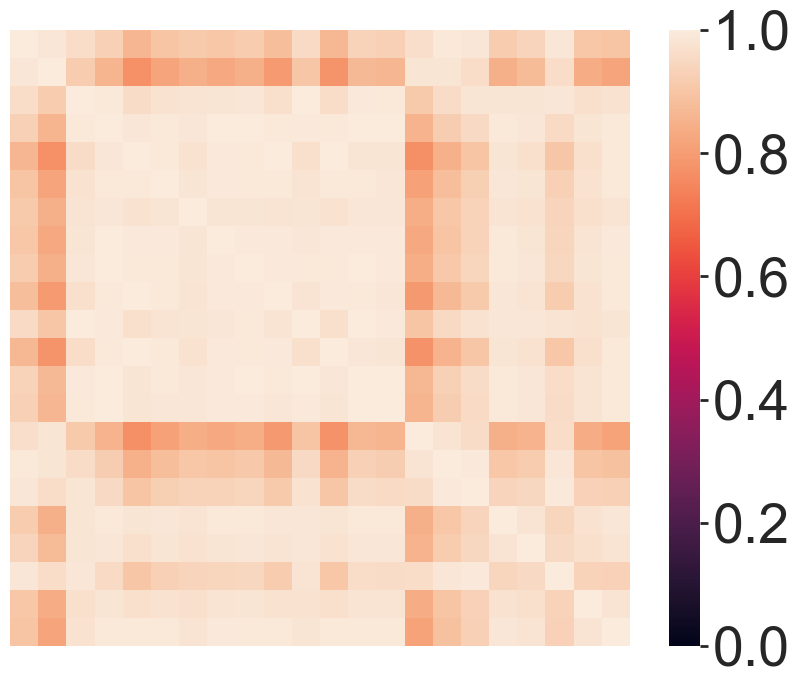}
  \caption{Spike2Vec}
\end{subfigure}%
\begin{subfigure}{.25\textwidth}
  \centering
  \includegraphics[scale = 0.120] {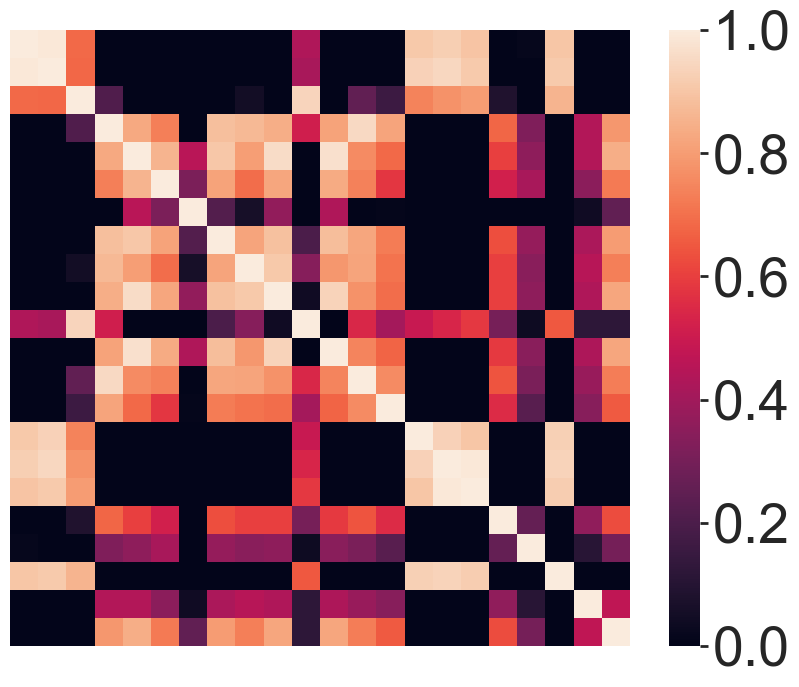}
  \caption{Hyperboloid Model}
\end{subfigure}
\caption{Heatmap for classes in \textbf{Spike7k dataset}. 
}
\label{fig_heat_map_7000}
\end{figure}

The classification results for the Human DNA dataset are summarized in Table~\ref{tbl_results_classification_human_dna}, representing the average performance across multiple evaluation metrics. Notably, the proposed hyperboloid model demonstrates superior performance compared to all other methods, achieving higher average accuracy, recall, weighted F1, macro F1, and ROC-AUC scores. These results highlight the effectiveness of the proposed kernel method in distinguishing between biological sequences, surpassing the performance of the baselines.

An interesting observation is the poor performance of the pre-trained Protein Bert model on the Human DNA dataset compared to its performance on the other two datasets. This discrepancy can be attributed to the fact that Protein Bert is specifically trained on protein sequence data, while the Human DNA dataset consists of nucleotide sequences. Hence the Protein Bert failed to generalize effectively in this case. In contrast, the proposed method exhibits superior performance compared to the baseline models, emphasizing their effectiveness in handling the Human DNA dataset.

\begin{table}[h!]
\centering
\resizebox{0.9\textwidth}{!}{
 \begin{tabular}{@{\extracolsep{6pt}}p{1.5cm}lp{1.1cm}p{1.1cm}p{1.1cm}p{1.3cm}p{1.3cm}p{1.1cm}p{1.7cm}}
    \toprule
        \multirow{2}{*}{Embeddings} & \multirow{2}{*}{Algo.} & \multirow{2}{*}{Acc. $\uparrow$} & \multirow{2}{*}{Prec. $\uparrow$} & \multirow{2}{*}{Recall $\uparrow$} & \multirow{2}{1.7cm}{F1 (Weig.) $\uparrow$} & \multirow{2}{1.7cm}{F1 (Macro) $\uparrow$} & \multirow{2}{1.2cm}{ROC AUC $\uparrow$} & Train Time (sec.) $\downarrow$\\
        \midrule \midrule	
        \multirow{7}{1.2cm}{PWM2Vec}
        & SVM &  0.302 & 0.241 & 0.302 & 0.165 & 0.091 & 0.505 & 10011.3 \\
         & NB &  0.084 & \underline{0.442} & 0.084 & 0.063 & 0.066 & \underline{0.511} & 4.565 \\
         & MLP &  \underline{0.310} & 0.350 & \underline{0.310} & 0.175 & 0.107 & 0.510 & 320.555 \\
         & KNN &  0.121 & 0.337 & 0.121 & 0.093 & 0.077 & 0.509 & 2.193 \\
         & RF &  0.309 & 0.332 & 0.309 & \underline{0.181} & 0.110 & 0.510 & 65.250 \\
         & LR & 0.304 & 0.257 & 0.304 & 0.167 & 0.094 & 0.506 & 23.651 \\
         & DT &  0.306 & 0.284 & 0.306 & \underline{0.181} & \underline{0.111} & 0.509 & \underline{1.861} \\
         \cmidrule{2-9} 
        \multirow{7}{1.5cm}{String Kernel}
        & SVM  &  0.618 & 0.617 & 0.618 & 0.613 & 0.588 & 0.753 & 39.791 \\
         & NB   &  0.338 & 0.452 & 0.338 & 0.347 & 0.333 & 0.617 & \underline{0.276} \\
         & MLP  & 0.597 & 0.595 & 0.597 & 0.593 & 0.549 & 0.737 & 331.068 \\
         & KNN  &  0.645 & 0.657 & 0.645 & 0.646 & 0.612 & 0.774 & 1.274 \\
         & RF   &  \underline{0.731} & \underline{0.776} & \underline{0.731} & \underline{0.729} & \underline{0.723} & \underline{0.808} & 12.673 \\
         & LR   &  0.571 & 0.570 & 0.571 & 0.558 & 0.532 & 0.716 & 2.995 \\
         & DT   & 0.630 & 0.631 & 0.630 & 0.630 & 0.598 & 0.767 & 2.682 \\
          \cmidrule{2-9} 
           \multirow{7}{1.2cm}{WDGRL}  
             & SVM & 0.318 & 0.101 & 0.318 & 0.154 & 0.069 & 0.500 & 0.751 \\
             & NB & 0.232 & 0.214 & 0.232 & 0.196 & 0.138 & 0.517 & \textbf{\underline{0.004}} \\
             & MLP &  0.326 & 0.286 & 0.326 & 0.263 & 0.186 & 0.535 & 8.613 \\
             & KNN & 0.317 & 0.317 & 0.317 & 0.315 & 0.266 & 0.574 & 0.092 \\
             & RF & \underline{0.453} & \underline{0.501} & \underline{0.453} & \underline{0.430} & \underline{0.389} & \underline{0.625} & 1.124 \\
             & LR & 0.323 & 0.279 & 0.323 & 0.177 & 0.095 & 0.507 & 0.041 \\
             & DT & 0.368 & 0.372 & 0.368 & 0.369 & 0.328 & 0.610 & 0.047 \\
         \cmidrule{2-9} 
            \multirow{7}{1.5cm}{Autoencoder}
             & SVM & 0.621 & 0.638 & 0.621 & 0.624 & 0.593 & 0.769 & 22.230 \\
             & NB &  0.260 & 0.426 & 0.260 & 0.247 & 0.268 & 0.583 & 0.287 \\
             & MLP & 0.621 & 0.624 & 0.621 & 0.620 & 0.578 & 0.756 & 111.809 \\
             & KNN & 0.565 & 0.577 & 0.565 & 0.568 & 0.547 & 0.732 & \underline{1.208} \\
             & RF & 0.689 & 0.738 & 0.689 & 0.683 & 0.668 & 0.774 & 20.131 \\
             & LR & \underline{0.692} & \underline{0.700} & \underline{0.692} & \underline{0.693} & \underline{0.672} & \underline{0.799} & 58.369 \\
             & DT & 0.543 & 0.546 & 0.543 & 0.543 & 0.515 & 0.718 & 10.616 \\
              \cmidrule{2-9} 
            \multirow{7}{1.5cm}{SeqVec}
             & SVM & 0.656 & 0.661 & 0.656 & 0.652 & 0.611 & \underline{0.791} & 0.891 \\
             & NB & 0.324 & 0.445 & 0.312 & 0.295 & 0.282 & 0.624 & \underline{0.036} \\
             & MLP & 0.657 & 0.633 & 0.653 & 0.646 & 0.616 & 0.783 & 12.432 \\
             & KNN & 0.592 & 0.606 & 0.592 & 0.591 & 0.552 & 0.717 & 0.571 \\
             & RF & 0.713 & \underline{0.724} & 0.701 & 0.702 & \underline{0.693} & 0.752 & 2.164 \\
             & LR & \underline{0.725} & 0.715 & \underline{0.726} & \underline{0.725} & 0.685 & 0.784 & 1.209 \\
             & DT & 0.586 & 0.553 & 0.585 & 0.577 & 0.557 & 0.736 & 0.24 \\
             \cmidrule{2-9} 
            \multirow{1}{1.9cm}{Protein Bert}
             & \_ & 0.542 & 0.580 & 0.542 & 0.514 & 0.447 & 0.675 & 58681.57 \\
              \cmidrule{2-9} 
            \multirow{7}{1.5cm}{Hyperboloid Model (ours)}
              & SVM & 0.508 & 0.441 & 0.508 & 0.458 & 0.357 & 0.642 & 8.440 \\
             & NB & 0.260 & 0.320 & 0.260 & 0.236 & 0.200 & 0.553 & \underline{0.101} \\
             & MLP & 0.574 & 0.580 & 0.574 & 0.575 & 0.544 & 0.736 & 4.265 \\
             & KNN & \textbf{0.739} & \textbf{0.748} & \textbf{0.739} & \textbf{0.740} & \textbf{0.724} & \textbf{0.836} & 2.339 \\
             & RF & 0.687 & 0.770 & 0.687 & 0.686 & 0.689 & 0.779 & 3.621 \\
             & LR & 0.488 & 0.437 & 0.488 & 0.429 & 0.338 & 0.627 & 0.790 \\
             & DT & 0.583 & 0.586 & 0.583 & 0.583 & 0.545 & 0.739 & 1.256 \\
         \bottomrule
         \end{tabular}
}
 \caption{Classification results (averaged over $5$ runs) on \textbf{Human DNA} dataset for different evaluation metrics. The best values are shown in bold. The best value for each embedding method is shown with the underline.}
    \label{tbl_results_classification_human_dna}
\end{table}

\subsection{Statistical Analysis}
To assess the statistical significance of the results, the student t-test was conducted using the average values and standard deviations (SD) from $5$ runs. The SD values for all metrics were found to be very small i.e.  $<0.002$. Hence, the majority of the computed p-values were less than 0.05, indicating the statistical significance of the results. 

\subsection{Inter-Class Embedding Interaction}
We utilize heat maps to analyze further whether our proposed kernel can identify different classes better. These maps are generated by first taking the average of the similarity values to compute a single value for each pair of classes and then computing the pairwise cosine similarity of different class's embeddings with one another. The heat map is further normalized between [0-1] to the identity pattern. The heatmaps for the baseline and its comparison with the proposed method embeddings are reported in Figure~\ref{fig_heat_map_5558}, ~\ref{fig_heat_map_4380} and ~\ref{fig_heat_map_7000}. We can observe that in the case of the data heatmap, the embeddings for the label are towards similar side. This eventually means it is difficult to distinguish between different classes due to high pairwise similarities among their vectors. 
On the other hand, we can observe that the pairwise similarity between different class embeddings is distinguishable for proposed method embeddings. This essentially means that the embeddings that belong to similar classes are highly similar to each other. In contrast, the embeddings for different classes are very different, indicating that the proposed methodology can accurately identify similar classes and different classes. 

The obtained results indicate that the proposed method consistently outperforms all baseline methods across different datasets in terms of average classification performance. This demonstrates the generalizability of the proposed approach, as it achieves higher predictive accuracy on both protein and nucleotide datasets. Furthermore, compared to the traditional String kernel method, the proposed approach exhibits superior performance, emphasizing its effectiveness in preserving information in the higher-dimensional space by capturing hierarchical and structural information. Additionally, the statistical significance of the results on real-world biological sequence datasets is demonstrated.

\section{Conclusion}\label{sec_conclusion}
In this study, we introduced a novel approach for the analysis of biological sequences by transforming their feature representation into the hyperboloid space. By leveraging this transformation, the proposed method preserved the hierarchical and structural information inherent in the sequences, addressing the limitations of conventional machine learning approaches operating in the Euclidean spaces.
The experimental evaluation showcased the efficacy of the proposed approach in capturing important sequence correlations and improving classification accuracy. We also provide theoretical detail regarding the important kernel properties.
Furthermore, the statistical significance of the results was confirmed through the student t-test.
Future research directions involve exploring the applications of the hyperboloid space in other areas of bioinformatics and investigating its performance on larger and more diverse biological datasets. 

\section{Funding}
Not Applicable

\section{Conflicts of interest/Competing interests}
\textcolor{black}{The authors do not have any conflict of interest}

\section{Ethics approval}
Not Applicable

\section{Consent to participate}
Not Applicable

\section{Consent for publication}
Not Applicable

\section{Availability of data and material}
Available on request

\section{Code availability}
Available on request

\section{Authors' contributions}
S. Ali and H. Mansoor worked on algorithm design. S. Ali performed experiments. H. Mansoor worked on theoretical analysis. M. Patterson supervised the project. All authors wrote and reviewed the manuscript

\bibliography{references}


\end{document}